\documentclass[11pt]{article}
\usepackage{algorithm}
\usepackage{algpseudocode}
\usepackage{amsmath}
\usepackage{amssymb}
\usepackage{xcolor}
\usepackage{xspace}
\newcommand{\ours}{\textsc{TDP}\xspace}
\definecolor{mygr}{RGB}{0,128,0}
\definecolor{AvgBlue}{HTML}{3F57A6}
\newcommand{\COMMENTLLAMA}[1]{{\color{mygr} $\triangleright$ {#1}}}
\usepackage{float}
\usepackage{booktabs}
\usepackage{multirow}
\usepackage{array}
\usepackage{colortbl}
\usepackage{multicol}
\usepackage[most]{tcolorbox}
\tcbuselibrary{breakable, skins}
\newtcblisting{promptbox}[1]{
  listing only,
  breakable,
  colback=gray!5,
  colframe=black!70,
  colbacktitle=black!85,
  coltitle=white,
  fonttitle=\bfseries,
  title=#1,
  boxrule=0.8pt,
  arc=2pt,
  left=6pt,
  right=6pt,
  top=6pt,
  bottom=6pt,
  width=\textwidth,
  listing options={
    basicstyle=\ttfamily\footnotesize,
    breaklines=true,
    breakindent=0pt,
    columns=fullflexible,
    keepspaces=true,
    showstringspaces=false
  }
}
\usepackage[most]{tcolorbox}
\tcbuselibrary{breakable, skins}

\newtcblisting{casebox}[1]{
  listing only,
  breakable,
  colback=green!5!white,         
  colframe=green!25!black,       
  colbacktitle=green!35!black,   
  coltitle=white,        
  fonttitle=\bfseries,
  title=#1,
  boxrule=0.8pt,
  arc=3pt,                   
  left=6pt,
  right=6pt,
  top=6pt,
  bottom=6pt,
  width=\columnwidth,         
  listing options={
    basicstyle=\ttfamily\footnotesize,
    breaklines=true,
    breakindent=0pt,
    columns=fullflexible,
    keepspaces=true,
    showstringspaces=false
  }
}

\usepackage{acl}

\usepackage{times}
\usepackage{latexsym}

\usepackage[T1]{fontenc}

\usepackage[utf8]{inputenc}

\usepackage{microtype}

\usepackage{inconsolata}

\usepackage{graphicx}

\title{Beyond Entangled Planning: Task-Decoupled Planning for \\ Long-Horizon Agents}

\author{
    Yunfan Li$^{1,2}$,
    Bingbing Xu$^{1}$\thanks{\; Corresponding author.}, \\ 
    \textbf{Xueyun Tian}$^{1,2}$, 
    \textbf{Xiucheng Xu}$^{1,2}$, 
    \textbf{Huawei Shen}$^{1,2}$\\
        $^{1}$State Key Laboratory of AI Safety, Institute of Computing Technology, CAS, \\
        $^{2}$University of Chinese Academy of Sciences \\
        \texttt{\{liyunfan24s,xubingbing\}@ict.ac.cn}
}

\begin{document}
\maketitle
\begin{abstract}
Recent advances in large language models (LLMs) have enabled agents to autonomously execute complex, long-horizon tasks, yet planning remains a primary bottleneck for reliable task execution. Existing methods typically fall into two paradigms: step-wise planning, which is reactive but often short-sighted; and one-shot planning, which generates a complete plan upfront yet is brittle to execution errors. Crucially, both paradigms suffer from entangled contexts, where the agent must reason over a monolithic history spanning multiple sub-tasks. This entanglement increases cognitive load and lets local errors propagate across otherwise independent decisions, making recovery computationally expensive.
To address this, we propose Task-Decoupled Planning (\ours), a training-free framework that replaces entangled reasoning with task decoupling. TDP decomposes tasks into a directed acyclic graph (DAG) of sub-goals via a \textit{Supervisor}. Using a \textit{Planner} and \textit{Executor} with scoped contexts, TDP confines reasoning and replanning to the active sub-task. This isolation prevents error propagation and corrects deviations locally without disrupting the workflow. Results on TravelPlanner, ScienceWorld, and HotpotQA show that TDP outperforms strong baselines while reducing token consumption by up to 82\%, demonstrating that sub-task decoupling improves both robustness and efficiency for long-horizon agents.

\end{abstract}

\section{Introduction}

Recent advances in large language models (LLMs)~\cite{Achiam2023GPT4TR,DeepSeekAI2024DeepSeekV3TR,Yang2024Qwen25TR} have enabled the emergence of LLM-based agents capable of autonomously performing complex, long-horizon tasks requiring interdependent decisions and environment interactions. Beyond language understanding and reasoning, successful task execution critically depends on an agent's planning capability to structure objectives, reason over alternative courses of action, and adapt decisions based on environmental feedback~\cite{Zhang2024ASO,Huang2024UnderstandingTP,Aghzal2025ASO,Hu2024AgentGenEP}. As tasks grow longer and more compositional, effective planning becomes a central bottleneck for robust agent behavior~\cite{Gao2025ASO}.

Existing planning methods for LLM-based agents can be broadly divided into step-wise planning and one-shot planning, which differ primarily in how planning and execution are interleaved. Step-wise approaches, such as ReAct~\cite{Yao2022ReActSR} and its extensions ReCode~\cite{Yu2025ReCodeUP} which further regard thinking itself as a high level action, interleave reasoning and acting at a fine granularity, enabling strong reactivity to feedback but often suffering from short-sighted decisions in long-horizon tasks. In contrast, one-shot planning methods generate high-level plans before execution and follow them throughout the task, sometimes augmented with intermediate structures or limited replanning. While these methods provide a global view, they are often brittle to execution errors and environmental uncertainty.
\begin{figure*}[!t]
    \centering
    \includegraphics[width=\linewidth]{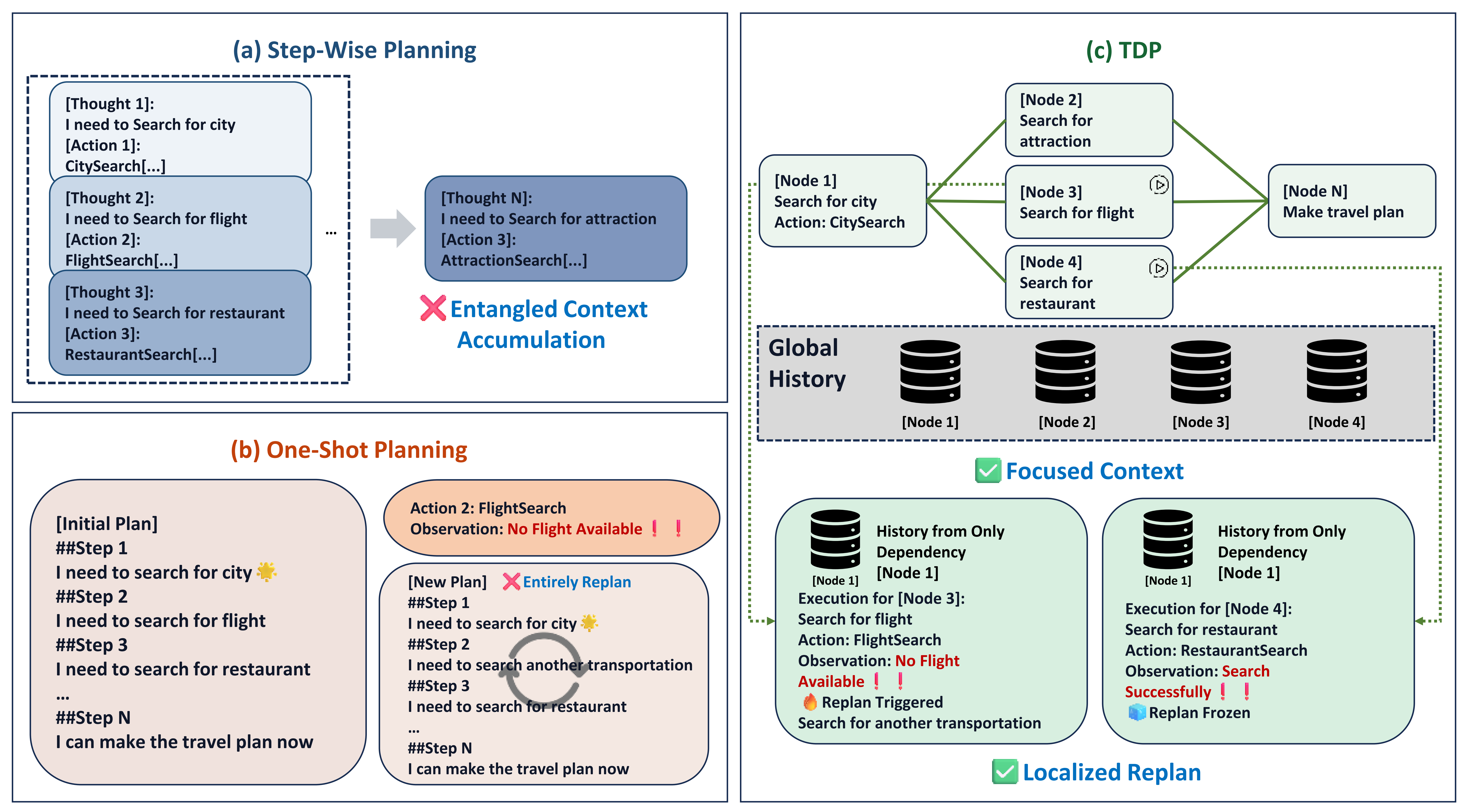}
    \caption{Comparison between step-wise planning (a), on-shot planning (b) and \ours (c) on a TravelPlanner task.}
    \label{fig.1}
\end{figure*}

Despite their differences, these approaches share a common design pattern: they focus on adjusting planning granularity, while treating the agent's internal workflow as a largely monolithic process. Planning is typically performed over long, entangled contexts that span multiple sub-tasks, and execution errors in one part of the task often trigger replanning over unrelated decisions. A single, growing execution history that mixes information across multiple sub-tasks makes planning process entangled. This coupling leads to two fundamental limitations.
First, planning over long histories that span multiple sub-tasks imposes substantial cognitive load on the model, increasing the difficulty of maintaining coherent and accurate reasoning throughout the planning process.
Second, execution failures incur high costs, as local errors may propagate to independent decisions, forcing unnecessary replanning and degrading both efficiency and robustness.
At their core, these issues arise from tight coupling between sub-tasks in planning and execution, rather than from the choice of planning granularity itself.

In this work, we propose Task-Decoupled Planning  (\ours), a training-free modular planning framework that addresses these limitations via explicit task decoupling. Instead of refining planning granularity in a monolithic workflow, it decouples an agent's decision process, with the core idea of localizing context, decisions, and error correction at the sub-task level. Concretely, \ours comprises a Supervisor that decomposes the original task into a directed acyclic graph (DAG) of sub-tasks, a Planner that generates sub-task-level plans conditioned solely on node-relevant context, and an Executor that translates plans into executable actions and interacts with the environment. Execution feedback triggers local replanning within a single node without propagating to independent sub-tasks. This isolation of decision dependencies and limited replanning scope ultimately reduces error propagation, lowers reasoning overhead, and enhances robustness in complex, long-horizon task execution. Figure~\ref{fig.1} contrasts existing paradigms and instantiates a task-decoupled paradigm in \ours.

We evaluate \ours on three challenging benchmarks, TravelPlanner~\cite{Xie2024TravelPlannerAB}, ScienceWorld~\cite{Wang2022ScienceWorldIY} and HotpotQA~\cite{Yang2018HotpotQAAD}, focusing on requirements for both fine-grained adaptability to real-time environmental interactions and long-horizon, global planning capabilities. Results show that \ours consistently achieves superior or competitive performance against strong baselines, effectively satisfying complex constraints and accumulating higher rewards. Notably, our framework attains these gains with significantly reduced token consumption compared to prior planning methods, demonstrating that the proposed sub-task decoupling strategy offers a highly efficient solution without compromising reasoning depth or task success.

Our main contributions are as follows:
\begin{itemize}
    \item \textbf{Decoupled Planning Paradigm}: We propose a novel framework that shifts from merely adjusting planning granularity to an explicit task-decoupling architecture, effectively separating task topology from execution logic.

    \item \textbf{The \ours Framework}: We introduce a modular design featuring decouple-based task abstraction, which localizes reasoning context and isolates off-plan outcomes during execution, mitigating error propagation and reasoning drift in long-horizon tasks.
    \item \textbf{Extensive Empirical Validation}: We evaluate \ours on three diverse benchmarks, TravelPlanner, ScienceWorld, and HotpotQA, where it outperforms state-of-the-art baselines in both robustness and efficiency.
\end{itemize}

\section{Related Work}
Automated planning and decision making are core capability of autonomous agents and requires substantial reasoning and decision-making~\cite{Huang2024UnderstandingTP}. Classical approaches largely relied on symbolic planning with formal models such as PDDL~\cite{PDDL} or sequential decision-making~\cite{1995Managing} via policy learning~\cite{He2015DeepRL,Yao2020KeepCA}. With the recent advances in large language models (LLMs), serving as reasoning and control modules of agents, LLMs have made these classical ideas revisited under a new paradigm~\cite{Huang2024UnderstandingTP,Song2024TrialAE,Chu2023NavigateTE}. However, LLMs remain imperfect as standalone planners, motivating structured frameworks that collaborate with classical methods~\cite{Aghzal2025ASO}.
\begin{figure*}[!ht]
    \centering
    \includegraphics[width=\linewidth]{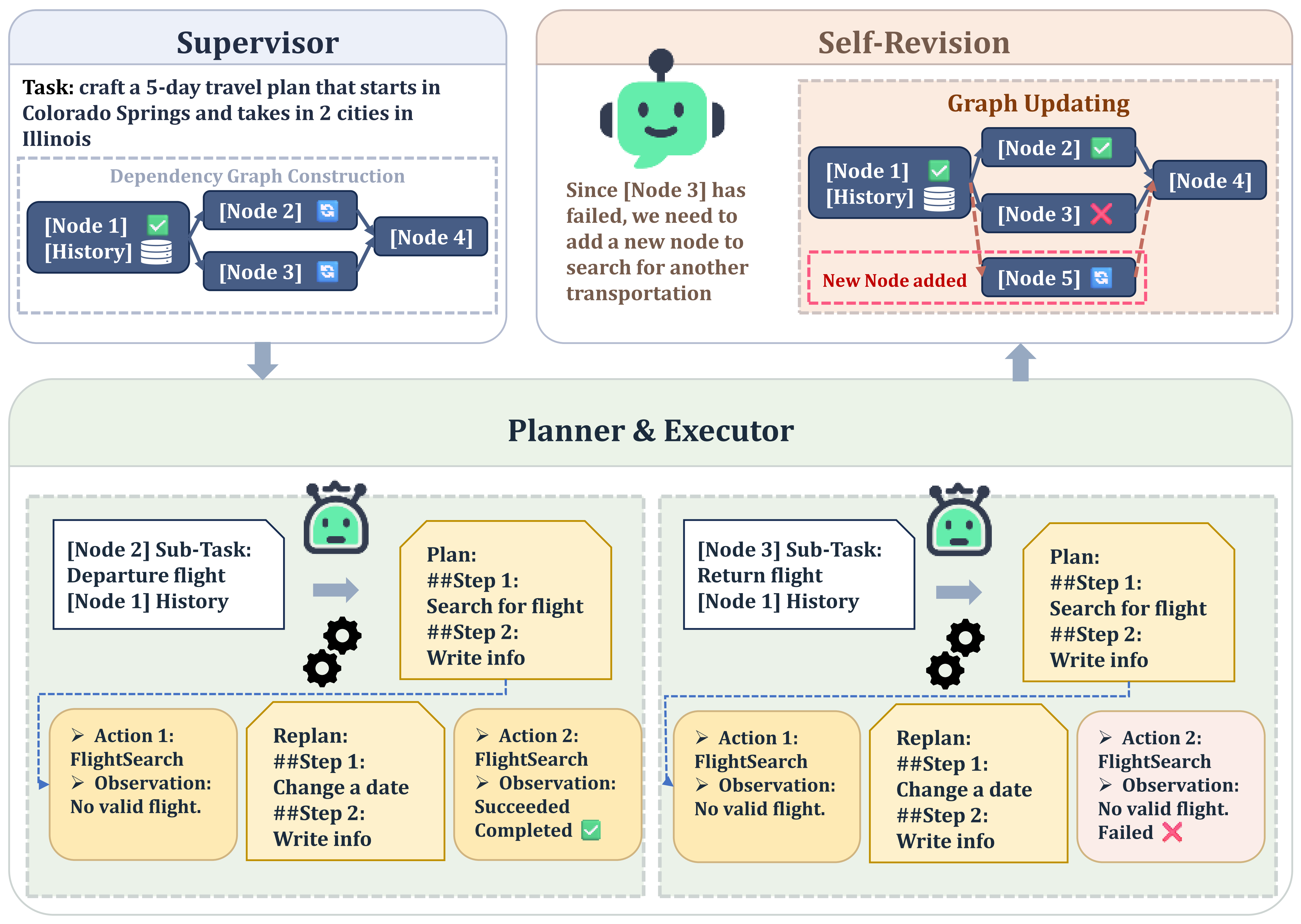}
    \caption{Overview of \ours. The Supervisor decomposes the task into a dependency graph; the Planner \& Executor solve each decoupled sub-task node; Self-Revision updates the graph after execution.}
    \label{fig:method}
\end{figure*}
\paragraph{LLM-based Agent Planning}
Early paradigms treat planning as prompted reasoning, where CoT~\cite{Wang2022SelfConsistencyIC}, ToT~\cite{yao2023tree}, and GoT~\cite{GoT} elicit structured reasoning paths. To enable interaction, ReAct~\cite{Yao2022ReActSR} interleaves reasoning with actions, while CodeAct~\cite{wang2024executable} and ReCode~\cite{Yu2025ReCodeUP} unify them via code or flexible representations. AgentFlow~\cite{Li2025IntheFlowAS} further structures this via modular loops. Conversely, global planning methods like Pre-Act~\cite{Rawat2025PreActMP} and Plan-and-Act~\cite{Erdogan2025PlanandActIP} explicitly separate multi-step planning from acting, with GoalAct~\cite{Chen2025EnhancingLA} and HiPlan~\cite{Li2025HiPlanHP} adopting hierarchical structures. Finally, AdaPlanner~\cite{NEURIPS2023_b5c8c1c1} and MPO~\cite{Xiong2025MPOBL} incorporate feedback-driven mechanisms for dynamic plan refinement.

\paragraph{Task Decomposition} 
\paragraph{Task Decomposition.} 
Task decomposition breaks a target into sub-tasks~\cite{2000Cognitive} and solves them in a divide-and-conquer manner. Existing approaches fall into \textit{agentic execution planning} and \textit{prompt-based reasoning enhancement}. 
For execution planning, HuggingGPT~\cite{NEURIPS2023_77c33e6a} uses LLMs as a controller to decompose user requests for expert models, while ProgPrompt~\cite{Singh2022ProgPromptGS} translates natural language descriptions into executable coding problems. 
Regarding reasoning enhancement, Decomposed Prompting~\cite{Khot2022DecomposedPA} breaks tasks into prompt-defined sub-tasks managed by specialized handlers, and Least-to-Most~\cite{Zhou2022LeasttoMostPE} decomposes complex queries into sequential subproblems to improve generalization.

\ours combines task decomposition with a modular agent design to enable sub-task-level decoupled planning and execution. Departing from methods that rely on fixed granularity or implicit task structures, this decoupling facilitates context isolation and strictly confines replanning to affected regions. 

\section{Method}
\label{sec:method}

To achieve focused context and localized recovery in long-horizon execution, we propose \textsc{\ours}, a modular agent framework that enforces \textbf{sub-task decoupling} by design. 
\textsc{\ours} separates global task structuring from node-level decision making: a \textbf{Supervisor} decomposes the task into a dependency graph and supervises execution (Section~\ref{sec:perception}), while a \textbf{Planner--Executor} pair plans and acts using only the context of the currently dispatched sub-task (Section~\ref{sec:local}). 
To preserve long-horizon coherence, a \textbf{Self-Revision} module updates the dependency graph and refines downstream sub-task specifications as execution unfolds (Section~\ref{sec:self-revision}), ensuring that deviations are handled locally while maintaining end-goal feasibility.

\subsection{Motivation}
\label{sec:motivation}

Most LLM-based planning methods implicitly operate as a monolithic workflow, reasoning over a single growing history that mixes states, decisions, and evidence from multiple sub-tasks. As horizons extend, this entangled planning makes it difficult to keep the model focused on the active sub-task, and local off-plan outcomes can trigger broad replanning over causally unrelated parts of the task.

To address this, we advocate explicit sub-task decoupling: each decision operates on a scoped context, and recovery is confined to the minimal affected scope. Guided by this principle, \ours separates global task structuring from node-level planning and execution, enabling localized replanning while preserving long-horizon coherence.

\subsection{Supervisor: Global Decomposition and Dependency-Aware Scheduling}
\label{sec:perception}

To provide a globally coherent structure while allowing independent execution, the \textbf{Supervisor} decomposes the overall objective into semantically coherent sub-tasks and infers prerequisite relations among them. This produces a directed dependency graph, where each node represents a decoupled sub-task with a standalone specification, such as two independent nodes with "search for flights" and "search for attractions".

To execute the graph under sub-task decoupling, the Supervisor performs topological scheduling and repeatedly selects \emph{ready} nodes whose prerequisites are satisfied. For each selected node, it constructs a \textbf{node-scoped context} for the sub-task by aggregating the node specification with the information retrieved from its predecessor dependencies, and then dispatches the node as an independent unit to downstream modules.This protocol anchors reasoning and recovery to the active sub-task, improving context focus and reducing interference from causally unrelated parts of the task.

To sustain long-horizon progress, the Supervisor also monitors execution at every step. After each environment interaction, it updates the status of the active node as \texttt{needs\_more\_steps}, \texttt{completed}, or \texttt{failed} based on the latest observation. When the node remains ongoing, it may additionally provide brief, macro-level guidance to steer downstream modules toward effective next steps.

\subsection{Planner and Executor: Local Planning, Acting, and Local Repair}
\label{sec:local}
To enforce context focus and confine recovery to a single sub-task, the Planner and Executor operate under strict locality after a node is dispatched. Each module receives (i) the current node specification and (ii) a compact, node-scoped context that contains only the prerequisite-node outcomes and the interaction trace accumulated while executing the current node (actions and observations). They never consume the full global execution history and sub-task decoupling is realized by this explicit context scoping.

Given a dispatched node, the \textbf{Planner} produces a structured high-level plan that enumerates intermediate objectives for completing the node. The \textbf{Executor} then follows the plan to generate concrete actions and interact with the environment step by step, iteratively incorporating returned observations into the ongoing execution trace.

When observations substantially conflict with the expected progress of the current plan, the system triggers \textbf{node-local replanning} to repair the trajectory without expanding scope. Replanning is restricted to the active node: only its plan and local execution trace are revised, while completed prerequisite nodes and causally independent nodes remain unchanged.

\subsection{Self-Revision: Graph Maintenance and Specification Refinement}
\label{sec:self-revision}

To keep the global task representation consistent with the evolving execution state under sub-task decoupling, we perform \textbf{Self-Revision} after each batch of ready nodes finishes. This step re-evaluates global progress and repairs the task dependency graph when needed.

Concretely, Self-Revision checks whether the outcomes of completed nodes violate the assumptions encoded in the current graph, including failure cases where the remaining dependencies can no longer support the overall objective. When such discrepancies arise, it revises the specifications of unfinished nodes and, if necessary, adds or removes nodes to restore a feasible dependency structure under the updated state.

Beyond failure repair, Self-Revision also consolidates newly acquired information from completed nodes for downstream use. When results resolve ambiguities or tighten constraints, it propagates them into the specifications of dependent nodes, making subsequent goals more explicit and concise. As a result, the next round of dispatched nodes receives better-scoped objectives and requires less case-by-case interpretation during local planning.

Figure~\ref{fig:method} overviews \ours. By decoupling sub-tasks and restricting downstream modules to node-scoped context, \ours anchors planning and execution to the active node, achieving context focus and confining deviations to node-local recovery. This design reduces cross-sub-task interference and avoids unnecessary global replanning, enabling reliable long-horizon progress across iterative rounds. Algorithm~\ref{alg:modular-agent} details the full workflow.

\begin{algorithm}[!h]
\caption{Task Loop with Global Supervision and Decoupled Planning and Execution}
\label{alg:modular-agent}
\begin{algorithmic}[1]
\Require Task instruction $\mathcal{T}$, environment $\mathcal{E}$, Supervisor $\mathbf{Sup}$, Planner $\mathbf{P}$, Executor $\mathbf{Ex}$, max steps $s_{max}$
\Ensure Task completion or termination status $\mathcal{R}$
\State Initialize global state $\mathcal{G}_0$, $steps \leftarrow 0$
\State $\mathcal{G} \leftarrow \mathbf{Sup}.\text{construct}(\mathcal{T}, \mathcal{E}, \mathcal{G}_0)$
\State\COMMENTLLAMA{Construct initial task DAG}
\While{\textbf{not} $\mathcal{T}.\text{done}$ \textbf{and} $steps \leq s_{max}$}
    \State $\mathcal{N} \leftarrow \text{GetExecutableNodes}(\mathcal{G})$
    \State\COMMENTLLAMA{Get executable (ready) nodes}
    \ForAll{node $n_i \in \mathcal{N}$}
        \State $\pi_i \leftarrow \mathbf{P}(n_i)$
        \State\COMMENTLLAMA{Generate high-level plan for node $n_i$}
        \State $\mathcal{H} \leftarrow \varnothing$
        \While{\textbf{not} $n_i.\text{done}$}
            \State $a \leftarrow \mathbf{Ex}(\pi_i, \mathcal{H})$
            \State $o \leftarrow \mathcal{E}.\textbf{Step}(a)$
            \State\COMMENTLLAMA{Execute action}
            \State $\langle d_i, r_i, \mathcal{H} \rangle \leftarrow \mathbf{Sup}.\text{eval}(n_i, o, \mathcal{H})$
            \State $n_i.\text{done} \leftarrow d_i,\quad n_i.\text{replan} \leftarrow r_i$
            \State\COMMENTLLAMA{Evaluate node status}
            \If{$n_i.\text{done}$}
                \State \textbf{break}
            \EndIf
            \If{$n_i.\text{replan}$}
                \State $\pi_i \leftarrow \mathbf{P}.\text{refine}(n_i, o, \mathcal{H})$
                \State\COMMENTLLAMA{Localized replanning}
            \EndIf
            \State $steps \leftarrow steps + 1$
            \If{$steps > s_{max}$}
                \State \textbf{break}
            \EndIf
        \EndWhile
        \If{$steps > s_{max}$}
            \State \textbf{break}
        \EndIf
    \EndFor
    \State $\mathcal{G} \leftarrow \mathbf{Sup}.\text{update}(\mathcal{G})$
    \State\COMMENTLLAMA{Self-revision: refine or expand DAG}
\EndWhile
\State \Return \textbf{Completed or Terminated}$(\mathcal{G})$
\end{algorithmic}
\end{algorithm}

\begin{table*}[!h]
\centering
\small
\setlength{\tabcolsep}{4pt}
\resizebox{\textwidth}{!}{
\begin{tabular}{l ccccccc ccc c}
\toprule
\multirow{2}{*}{Method}
& \multicolumn{7}{c}{\textbf{TravelPlanner}}
& \multicolumn{3}{c}{\textbf{HotpotQA}}
& \multicolumn{1}{c}{\textbf{ScienceWorld}} \\
\cmidrule(lr){2-8} \cmidrule(lr){9-11} \cmidrule(lr){12-12}
& Delivery $\uparrow$
& CS Micro $\uparrow$
& CS Macro $\uparrow$
& HC Micro $\uparrow$
& HC Macro $\uparrow$
& Final Pass $\uparrow$
& \textcolor{AvgBlue}{\textbf{Avg$\uparrow$}} 
& Accuracy $\uparrow$
& Deli. Acc. $\uparrow$
& \textcolor{AvgBlue}{\textbf{Avg$\uparrow$}}
& Reward $\uparrow$ \\
\midrule
\rowcolor[HTML]{E2F0D9}\multicolumn{12}{c}{\textbf{DeepSeek-V3.2}} \\
ReAct
& \textbf{100\%} & \underline{60.52\%} & 0.83\% & \underline{15\%} & 2.50\% & 0\% & \textcolor{AvgBlue}{30\%}
& \underline{72\%} & 73.47\% & \textcolor{AvgBlue}{\underline{72.74\%}}
& \textbf{58.26} \\
CoT
& 91\% & \textbf{61.12\%} & \textbf{1.67\%} & 10\% & \underline{3.33\%} & 0.83\% & \textcolor{AvgBlue}{\underline{28\%}}
& 55\% & 75.34\% & \textcolor{AvgBlue}{65.17\%}
& 39.29 \\
Plan-and-Act
& 82.50\% & 55.18\% & 0.83\% & 10\% & \underline{3.33\%} & 0.83\% & \textcolor{AvgBlue}{25\%}
& 57\% & \underline{85.07\%} & \textcolor{AvgBlue}{71.04\%}
& 45.38 \\
\ours
& \underline{96.67\%} & 60.12\% & \underline{0.83\%} & \textbf{32.50\%} & \textbf{10.83\%} & \textbf{0.83\%} & \textcolor{AvgBlue}{\textbf{34\%}}
& \textbf{73\%} & \textbf{85.88\%} & \textcolor{AvgBlue}{\textbf{79.44\%}}
& \underline{56.17} \\
\midrule
\rowcolor[HTML]{E2F0D9}\multicolumn{12}{c}{\textbf{GPT-4o}} \\
ReAct
& \textbf{95.83\%} & 53.80\% & 0\% & 15.00\% & 5.00\% & 0.00\% & \textcolor{AvgBlue}{\underline{28\%}}
& \textbf{73\%} & 73\% & \textcolor{AvgBlue}{\underline{73.00\%}}
& \underline{50.63} \\
CoT
& 75.83\% & 58.24\% & \underline{1.67\%} & 10.00\% & 3.33\% & \underline{1.67\%} & \textcolor{AvgBlue}{25\%}
& 61\% & 75.31\% & \textcolor{AvgBlue}{68.16\%}
& 38.75 \\
Plan-and-Act
& 59.17\% & \textbf{62.32\%} & \textbf{4.17\%} & \underline{20.00\%} & \underline{6.67\%} & \textbf{2.50\%} & \textcolor{AvgBlue}{26\%}
& 48\% & \underline{81.36\%} & \textcolor{AvgBlue}{64.68\%}
& 49.69 \\
\ours
& \underline{94.17\%} & \underline{59.29\%} & 0.00\% & \textbf{20.00\%} & \textbf{6.67\%} & 0.00\% & \textcolor{AvgBlue}{\textbf{30\%}}
& \underline{69\%} & \textbf{82.14\%} & \textcolor{AvgBlue}{\textbf{75.57\%}}
& \textbf{53.24} \\
\bottomrule
\end{tabular}}
\caption{
Main results on TravelPlanner, HotpotQA, and ScienceWorld. The optimal and suboptimal results in each section are marked in \textbf{bold} and \underline{underlined}, respectively. Average scores across TravelPlanner and HotpotQA metrics are highlighted in \textcolor{AvgBlue}{blue}.
}
\label{tab:results}
\end{table*}

\section{Experiment}
We evaluate \ours on three representative long-horizon benchmarks requiring multi-step planning with iterative feedback and compare against strong baselines under two base LLMs. Our experiments emphasize whether sub-task decoupling improves context focus and local error containment without sacrificing final task performance.

\subsection{Experimental Setup}
\paragraph{Benchmarks}
We conduct experiments on TravelPlanner, HotpotQA, and ScienceWorld, covering constraint-heavy tool planning, multi-hop reasoning, and closed-loop interactive control, respectively. HotpotQA requires multi-hop evidence aggregation and long-chain reasoning over retrieved documents. ScienceWorld is a text-based interactive environment where agents must adapt to step-by-step feedback. TravelPlanner tests multi-stage tool calls with globally coupled constraints and long-horizon planning under evolving contexts. Together, these benchmarks allow us to assess robustness across heterogeneous task structures and interaction patterns.

\paragraph{Evaluation Metrics}
For HotpotQA, we report Accuracy and Accurate Delivery (Deli.\ Acc.), which measures answer correctness conditioned on successful task completion. We adopt an LLM-as-a-judge~\cite{Li2024FromGT} evaluation protocol, using Qwen-MAX as the judge LLM to determine whether a delivered answer is correct.
For ScienceWorld, we report the environment-provided average reward, a dense reward ranging from 0 to 1 based on task progress and solution accuracy provided by the environment.
For TravelPlanner, we follow the official protocol and report Delivery, Commonsense/Hard Constraint (CS/HC) micro- and macro-pass rates, and Final Pass, where micro measures the proportion of satisfied constraints and macro/final require satisfying all constraints in the corresponding category. We additionally report the arithmetic mean of all metrics on TravelPlanner and HotpotQA as Avg, to summarize overall constraint satisfaction quality.

\paragraph{Baselines}
We compare against ReAct, which interleaves reasoning and acting step-by-step based on the accumulated history, CoT with an initial plan (CoT), which generates a one-shot plan and executes without replanning, and \textbf{Plan-and-Act}, which combines high-level planning with step-wise execution and replans based on feedback.

All methods are evaluated with two base models, DeepSeek-V3.2 and GPT-4o, using matched settings for fair comparison.
Detailed benchmark configurations and implementation details are provided in the appendix.

\subsection{Experiment Results}
\label{sec:results-performance}
Table~\ref{tab:results} shows the main results of our experiment on three benchmarks. Overall, \ours exhibits the most stable behavior across heterogeneous long-horizon regimes, remains competitive on benchmarks that favor step-wise control, consistently improving the quality of completed trajectories on tasks where global structure and constraint satisfaction matter. 
\begin{figure*}[!t]
    \centering
    \includegraphics[width=\linewidth]{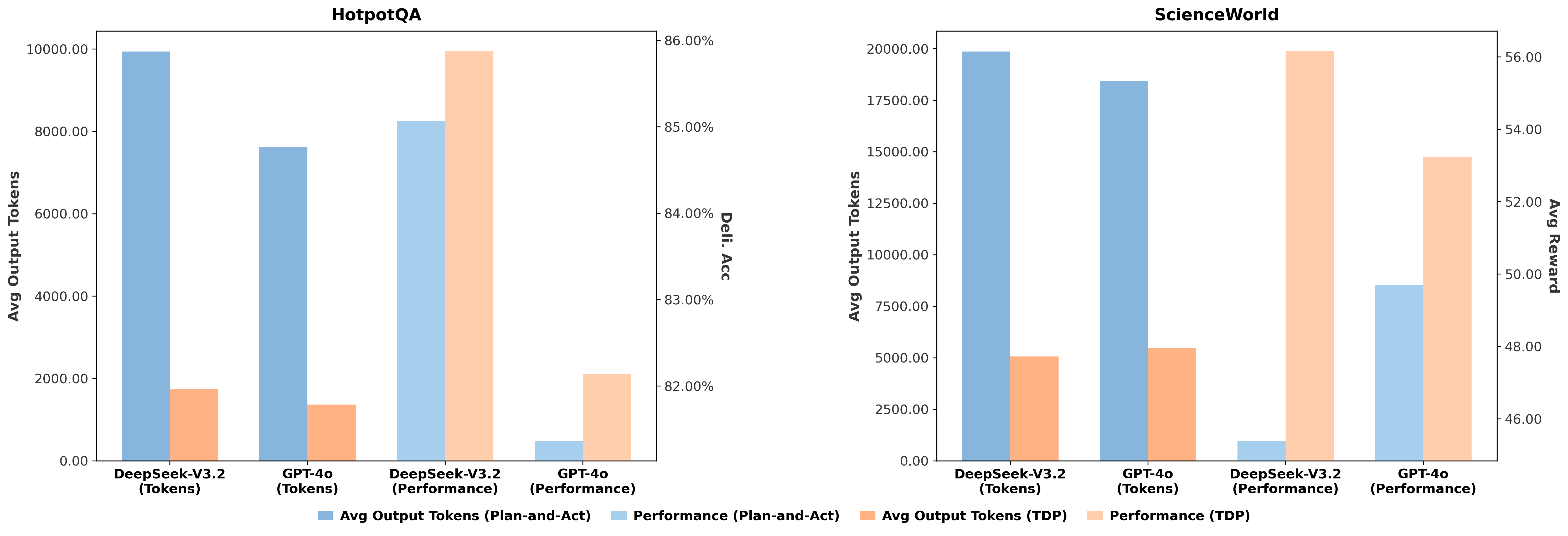}
    \caption{
    Cost comparison on HotpotQA and ScienceWorld: average output tokens (left axis) and performance (right axis; delivery accuracy / average reward) for Plan-and-Act vs TDP under DeepSeek-V3.2 and GPT-4o.
    }
    \label{fig:cost}
\end{figure*}

\paragraph{TravelPlanner}
On TravelPlanner, \ours achieves the best \textbf{Avg} score under both DeepSeek-V3.2 (34\%) and GPT-4o (30\%), while maintaining delivery rates comparable to ReAct and remaining among the top methods on constraint-related metrics. In particular, \ours achieves the highest micro pass rate on hard constraints (32.5\%) under DeepSeek-V3.2 and matches the best-performing methods on GPT-4o (20.0\%), and also attains the strongest hard-constraint macro performance under DeepSeek-V3.2 (10.83\%). \ours sustains a high delivery rate under the fixed step budget, while also maintaining strong performance on both the Commonsense and Hard constraint categories. TravelPlanner centers on constraint-heavy tool planning, where requirements can often be identified early but execution produces long, noisy traces that easily introduce interference across decisions. Sub-task decoupling mitigates this by scoping each sub-task to prerequisite information and its local interaction history, which keeps constraint-relevant details anchored to the sub-task scope and limits off-plan deviations to local recovery, supporting stable long-horizon execution.

\paragraph{HotpotQA}
On HotpotQA, \ours achieves the strongest delivered-answer quality under DeepSeek-V3.2 (85.88\%) and the best delivered accuracy among plan-based baselines under GPT-4o (82.14\%), while remaining competitive with ReAct in overall answer accuracy. Under both models, \ours consistently improves Accurate Delivery compared to ReAct and CoT, and it performs on par with or slightly above Plan-and-Act in delivered correctness, with especially clear advantages when the backbone model is DeepSeek-V3.2. At the same time, \ours preserves strong completion behavior under the same step limit, indicating that the quality gains are not driven by selective answering. HotpotQA requires multi-hop decomposition, evidence aggregation, and coherent synthesis without environment dynamics, so performance hinges on maintaining a clean reasoning state as the context grows. Step-wise methods often accumulate sprawling histories that increase the risk of drift, while one-shot planning can become brittle when early choices are suboptimal. Sub-task decoupling addresses this by structuring reasoning into locally focused units, limiting context carryover to prerequisite information and confining plan deviations to the active sub-task, which supports stable multi-hop reasoning and targeted corrections.

\paragraph{ScienceWorld}
On ScienceWorld, \ours achieves the best average reward under GPT-4o (53.24) and remains highly competitive with ReAct under DeepSeek-V3.2, while consistently outperforming CoT and Plan-and-Act across both models. This environment is partially observable and heavily driven by immediate feedback, so effective behavior depends on iterative interaction and rapid local adaptation. This feature determines that it naturally favors step-wise planning paradigms such as ReAct that tightly interleave acting and observation. Our framework preserves step-wise execution but organizes progress via decoupled sub-tasks, confining plan deviations to local recovery within the active sub-task and avoiding entanglement with unrelated parts of the trajectory, which supports robust closed-loop control over long horizons.

Overall, the results highlight a consistent pattern across regimes: baselines tend to excel in the task settings that match their planning granularity, whereas our method remains stable across all three benchmarks. Sub-task decoupling provides a common mechanism that fits heterogeneous demands, enabling context-focused reasoning for multi-hop problems, localized adaptation for interactive environments, and interference-resistant execution for constraint-heavy tool planning. Algorithm~\ref{alg:modular-agent} details the complete workflow used in our experiments.
\begin{figure*}
    \centering
    \includegraphics[width=\linewidth]{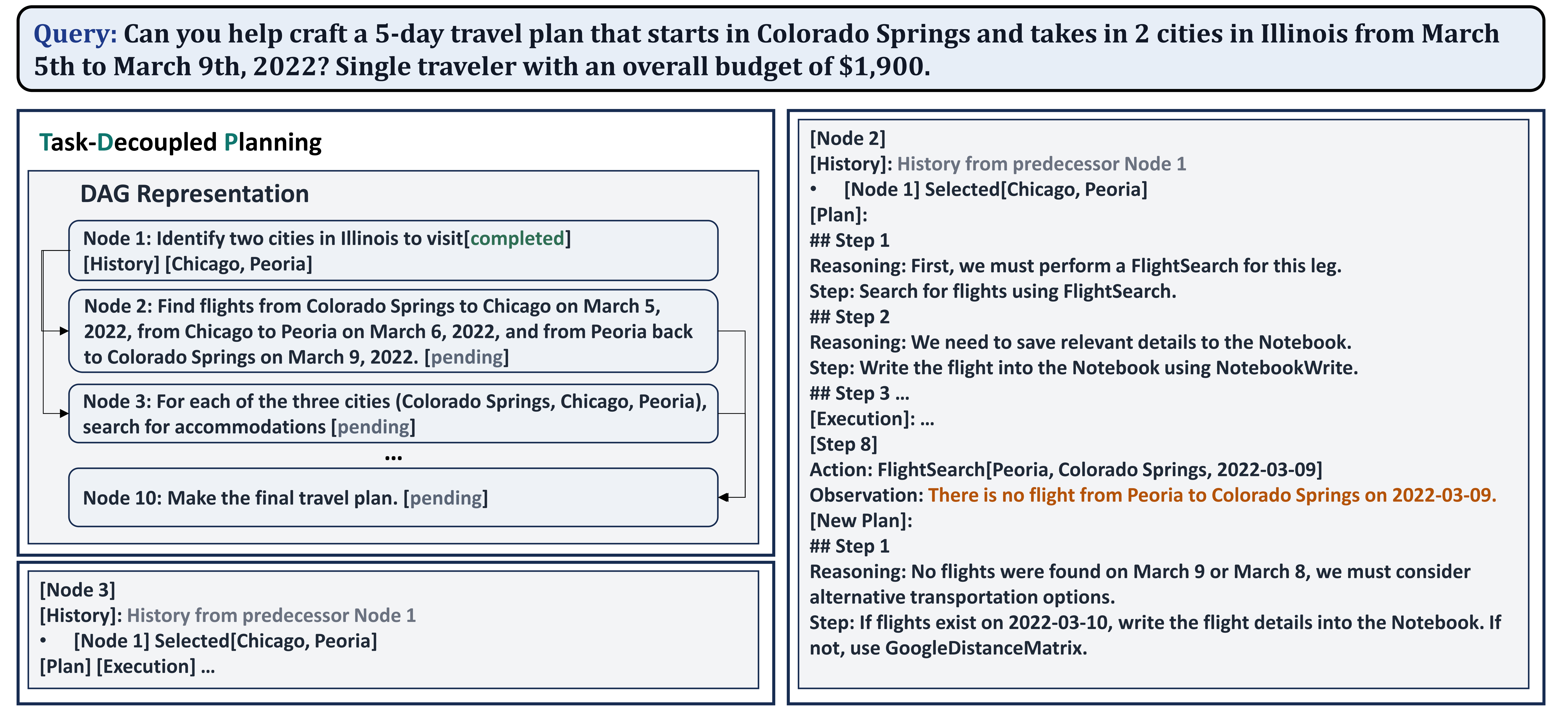}
    \caption{Illustration of \ours task decomposition, node-local planning and replanning and isolated context on a TravelPlanner case.}
    \label{fig:case}
\end{figure*}

\subsection{Cost Comparison}
\label{sec:results-cost}
Beyond task performance, we compare the overhead of handling off-plan outcomes by measuring average output tokens for \ours and Plan-and-Act on HotpotQA and ScienceWorld. The results are summarized in Figure~\ref{fig:cost}.

On HotpotQA, \ours uses 1,747.47 average output tokens with DeepSeek-V3.2 and 1,361.62 with GPT-4o, corresponding to an 82.4\% and 82.1\% reduction compared to Plan-and-Act. On ScienceWorld, \ours uses 5,060.57 average output tokens with DeepSeek-V3.2 and 5,461.65 with GPT-4o, reducing by 74.5\% and 70.4\% respectively. Across both benchmarks and both base models, \ours also delivers stronger task outcomes than Plan-and-Act, improving delivery accuracy on HotpotQA and increasing average reward on ScienceWorld.

These gains come from strict sub-task decoupling. Plan-and-Act resolves deviations by revising a global plan, which tends to expand the replanning scope and repeatedly re-justify decisions that are not causally affected by the current state. In contrast, \ours dispatches a node with node-scoped context and confines recovery to the active node, so replanning updates only the local plan and local execution trace while leaving independent sub-tasks intact. This prevents error propagation and avoids unnecessary global reconsideration, reducing token overhead while keeping decision making focused and stable over long horizons.

\subsection{Case Study}

To illustrate how \ours achieves context focus and localized replanning via sub-task decoupling, we present a TravelPlanner case study in Figure~\ref{fig:case}.

Since the user query does not specify which cities in Illinois to visit, the Supervisor first introduces an upstream sub-task (Node~1) to identify candidate cities. After Node~1 completes, downstream sub-tasks including Node~2 for collecting flight information and Node~3 for searching accommodations get ready. As visualized in Figure~\ref{fig:case}, these sub-tasks are dispatched independently with node-scoped contexts that include only prerequisite information and local execution history.

This decoupling prevents cross-sub-task context entanglement. Node~3 does not inherit the flight-search trace produced while executing Node~2, even though both depend on Node~1, local planning for accommodation search remains focused and free from irrelevant execution details.

Figure~\ref{fig:case} further demonstrates localized replanning. During the execution of Node~2, the agent observes that no return flight from Peoria to Colorado Springs is available on the planned date, triggering a plan deviation. \ours responds by replanning strictly within Node~2 and updating its local plan to consider alternative transportation options, while leaving other sub-tasks unchanged. This localized handling prevents error propagation across the DAG and allows independent sub-tasks to proceed without interruption.

\section{Conclusion}
In this work, we propose a sub-task decoupling framework for long-horizon tasks that separates global task decomposition and supervision from sub-task-level decoupled execution, moving beyond the entangled planning process. This framework design ensures task coherence while mitigating error propagation and improving context focus through strict scope constraints. Experiments across diverse benchmarks demonstrate the robustness and adaptability of \ours in complex settings. Ultimately, our results highlight that explicitly controlling planning scope is an effective design principle for balancing global awareness with local adaptive execution in complex environments.

\section*{Limitations}
While the proposed framework demonstrates strong stability, several aspects warrant further exploration. First, as the current approach prioritizes robust scope control, the granularity of sub-task decomposition may not always be strictly optimized for execution efficiency. Second, our evaluation primarily focuses on benchmarks with verifiable objectives; extending the framework to open-ended real-world tasks with subjective or implicit termination signals remains a promising avenue for future work.

\section*{Ethical Considerations}
The proposed method aims to improve agents' planning and execution capabilities in complex long-horizon tasks. Its primary contribution lies in enhancing efficiency and robustness through architectural design, rather than introducing new decision objectives or value preferences. As such, the method itself does not directly involve specific social value judgments or sensitive decision-making scenarios.

\bibliography{custom}
\newpage
\appendix

\section{Details of Benchmarks}
\label{sec:benchmark}
In this section we provide a detailed introduction or the three benchmarks.

\paragraph{HotpotQA}
HotpotQA is a widely used multi-hop question answering benchmark that requires agents to aggregate evidence from multiple supporting documents and perform a sequence of interdependent reasoning steps to derive the correct answer. Unlike single-hop QA tasks, HotpotQA emphasizes long-horizon reasoning, where intermediate information retrieval and reasoning decisions critically affect final performance.

In our experimental setup, we adopt an interactive formulation of HotpotQA that allows the agent to perform real-time information retrieval over Wikipedia, rather than relying on pre-retrieved documents. Three types of available actions are provided to the agents:
\begin{itemize}
    \item \texttt{Search[Keyword]}: Issues a search query to Wikipedia using the specified keyword. If a matching page exists, the environment returns the first paragraph of the corresponding article. If no exact match is found, a list of similar entities is returned to guide subsequent searches.
    \item \texttt{Lookup[Keyword]}: Retrieves additional information from the currently active Wikipedia page, defined as the most recently searched page. This action returns the next sentence in the page that contains the specified keyword, enabling fine-grained evidence extraction within a single document.
    \item \texttt{Finish[Answer]}: Terminates the interaction and outputs the agent's final answer in the form \texttt{Finish[your final answer here]}. This action is used when the agent determines that sufficient evidence has been gathered to answer the question.
\end{itemize}

\paragraph{ScienceWorld}
ScienceWorld is a text-based simulation environment designed around the execution of basic scientific experiments. In this environment, an agent interacts with a simulated world to perform experimental procedures, such as manipulating objects, observing environmental states, and measuring specific values, in order to complete predefined scientific tasks. Task completion typically requires applying elementary scientific knowledge through a sequence of environment interactions.

ScienceWorld provides dense rewards in the range of $[0,1]$ for each task, reflecting the degree of sub-goal completion achieved during execution.

\paragraph{TravelPlanner}
TravelPlanner is a tool-augmented, multi-turn benchmark designed to evaluate an agent's ability to complete complex travel planning tasks within multiple constraints through structured interaction with external information sources.

For a given query, the agent must gather relevant information, manage intermediate results, and finally produce a coherent and feasible travel plan. The benchmark integrates multiple heterogeneous tools and requires agents to coordinate information across different planning aspects, including transportation, daily meals, attractions, and accommodation for each day. Here are the tools provided with the agent to interact with the environment:
\begin{itemize}
    \item \texttt{FlightSearch[City, City, Date]}: Retrieves flight information between two cities on a specified date.
    \item \texttt{GoogleDistanceMatrix[City, City, Mode]}: Estimates distance, travel time, and cost between two different cities by a travel mode \textit{self-driving} or \textit{taxi}.
    \item \texttt{AccommodationSearch[City]}: Searches for accommodation options in a given city.
    \item \texttt{RestaurantSearch[City]}: Retrieves restaurant information for a specified city.
    \item \texttt{AttractionSearch[City]}: Retrieves tourist attraction information for a specified city.
    \item \texttt{CitySearch[State]}: Returns a list of cities within a given state from a predefined state set.
    \item \texttt{NotebookWrite[Content]}: Writes intermediate information to a persistent Notebook. This action is typically used after search operations to store key details such as flight numbers, dates, times, accommodation names, prices, or city lists.
    \item \texttt{MakePlan[Query]}: Generates a detailed travel plan based on the original user query and the information stored in the Notebook, using a dedicated planning tool.
\end{itemize}

For constraints, from the perspective of real world applications, TravelPlanner includes three types of them: Environment Constraint, Commonsense Constraint, and Hard Constraint. 
\section{Implementation Details}
On HotpotQA, We evaluate all methods on 100 instances sampled from the \texttt{hotpot\_dev\_fullwiki\_v1} split. The maximum number of execution steps is limited to 30 for all agents. On ScienceWorld, we evaluate on 100 instances from the official test split. Each episode is constrained to a maximum of 60 interaction steps. On TravelPlanner, we evaluate on the \texttt{validation} split, selecting 40 tasks from each difficulty level (easy, medium, and hard), resulting in a total of 120 tasks. The maximum number of execution steps is set to 50 for all methods.

\section{Prompt Templates}
\label{sec:appendix_prompt}
We provide detailed prompt templates for our method in this section. 6 prompt templates are used in our method: prompt for Supervisor to decompose the task and construct the initial DAG, prompt for Planner to generate plan for a single node, prompt for Executor to generate a single executable action, prompt for Supervisor to evaluate whether a node has been completed, prompt for Planner to replan for a node, and prompt for Supervisor to check whether and how to update the DAG.
\onecolumn
\begin{promptbox}{DAG Construction Prompt for Supervisor}

You are a High-level Supervisor for a task agent. Given a task, you need to decompose the task into subgoals and identify their dependencies, construct a DAG structure.
Then each node will be planned and executed by a separate Planner Module and an Executor Module.
Carefully decide the content of subgoal description, make sure the Planner Module and the Executor Module can clearly understand what to do at each node.
Planner will only see the information you provide and limited context, so concisely and clearly include what you know in the node description.
Each subgoal MUST be executable using Available Actions below. Only use actions from the list.

You will be provided with:
- **Task Description**: The task that you are required to complete.
- **Available Actions**: The available actions to use.

GUIDELINES:
1. **Complete task coverage**: Ensure your DAG covers ALL essential steps required by the task description. Make sure your DAG explicitly includes nodes for all required actions, including final actions to complete the task.
2. Your DAG MUST always contain at least one node that represents the final action to complete the task.
3. Create subgoals at semantic task level. Each subgoal represents a complete task unit, may require multiple actions, grouped as one semantic unit.
4. Do not create redundant subgoals.
5. When constructing the DAG, prioritize a direct path to task completion. Good decomposition should enable direct task completion once the necessary information and resources are gathered, avoiding unnecessary intermediate verification or redundant steps.

Task Description: {task_description}

Available actions:
{admissible_commands}

Your output should be in the following JSON format:
{{
  "subgoals": [
    {{
      "id": "node_1",
      "description": "Subgoal description (semantic level)",
      "dependencies": []
    }},
    {{
      "id": "node_2",
      "description": "Another subgoal (semantic level)",
      "dependencies": ["node_1"]
    }}
  ]
}}

Analyze and output the DAG structure in JSON:
\end{promptbox}

\begin{promptbox}{Plan Generation Prompt for Planner}
You are the Planner Module for a task. You will be given a subgoal (part of the overall task) and need to generate a structured, step-by-step plan to achieve this subgoal.
A separate Executor Module will execute your plan by converting each step into concrete actions. Ensure your plan is clear and actionable.

You will be provided with:
- **Task**: The overall task.
- **Current Subgoal**: The current subgoal to be achieved.
- **Available Actions**: The available actions to use.
- **History**: The history of actions and observations.

Focus ONLY on the subgoal above. Do NOT plan for other parts of the task. Check history to avoid redundant actions.

## Expected Output Format
Your plan should be structured as a numbered list, starting with '## Step 1'. Each step must follow this format:
## Step N
Reasoning: <Your reasoning explaining why this step is needed and how it contributes to the subgoal>
Step: <A concise description of what needs to be accomplished, specifying what to interact with or what action to take>

Guidelines:
- Focus on high-level goals while being specific about objects, locations, or parameters. Be explicit rather than vague.
- Describe WHAT needs to be accomplished, not HOW (the Executor will handle action execution).
- Keep each step concise (1-2 sentences max).
- Prioritize utilizing information already available in the history. Before planning new actions, check if the history already contains the needed information for this subgoal.
- Reduce redundant actions: Only include actions in your plan when they are essential for the subgoal.
- If the subgoal is to complete the task, directly use the final action.
- Keep efficient, design a plan that can be executed in fewer steps as possible.

Task: {task_description}
Current Subgoal: {nodes_description}
Available actions:
{admissible_commands}

History: {history}

Your response:
\end{promptbox}

\begin{promptbox}{Executor Prompt}
You are the Executor Module for a task. You will be given a structured plan and need to generate ONE immediate action to execute in the current step.
Your goal is to follow the plan, use the history to identify where you are in the plan, and output the next action to progress toward completing the subgoal.

You will be provided with:
- **Task Description**: The overall task.
- **Current Subgoal**: The current subgoal to be achieved.
- **Current Plan**: The current plan that you need to execute.
- **Guidance from the Supervisor**: The guidance from the Supervisor that provides hints for the next action to take.
- **Available Actions**: The available actions to use.
- **History**: The history of actions and observations.

EXECUTION Guidelines:
1. **Plan tracking**: Review the History to identify which steps have already been executed, then determine the next unexecuted step from the Current Plan.
2. **Action selection**
3. **Guidance priority**: If guidance is provided, prioritize it when choosing the next action, but still consider the plan context.

Task: {task_description}
Current Subgoal: {subgoal}
Current Plan: {plan}
Guidance from the Supervisor (may be "None" or empty): {guidance}

Available actions:
{admissible_commands}

History: {history}

Output ONLY the action string, nothing else.

Your response:
\end{promptbox}

\begin{promptbox}{Evaluate Execution Result Prompt for Supervisor}
You are a Supervisor evaluating whether a generic subgoal has been completed based on observation and action history, and determine if replanning is necessary.

Task Description: {task_description}
Current Subgoal: {subgoal}
Current Plan: {current_plan}
Available Actions: {admissible_commands}

History:
{history}

EVALUATION:
- Examine History to identify successfully executed actions
- Determine if any successful action has completed the current subgoal
- Check if observations from successful actions indicate subgoal requirements are met
- Compare execution progress with Current Plan to determine if plan is still valid
- When you judge that the status should be "needs_more_steps" and need_replan is false, you MUST also generate 1-2 short sentences of concrete guidance for the next action and put them into the "reason" field. This guidance will be provided to the Executor as a "guidance" field in the next step.

MINIMIZE replanning. Set need_replan to true ONLY if:
- Plan is clearly not working (actions failing repeatedly, environment state doesn't match plan assumptions)
- Critical obstacle that plan cannot handle
- Plan missing essential steps that are now apparent

Your response MUST be valid JSON:
{{
  "status": "<completed|failed|needs_more_steps>",
  "reason": "<1-2 sentences: guidance for the next step if status is 'needs_more_steps' and need_replan is false / brief explanation for the 'failed' status or the 'needs_more_steps' status if need_replan is true / null otherwise>",
  "need_replan": <true or false>
}}
\end{promptbox}

\begin{promptbox}{Replanning Prompt for Planner}
You are the Planner Module for a task. After the Executor executes an action, you need to evaluate whether the current plan should continue or be revised based on the Supervisor's assessment.
The Supervisor has evaluated the execution and provided a reason. If the reason indicates a fundamental problem with the plan that requires replanning, generate a new plan.

You will be provided with:
- **Task**: The overall task.
- **Current Subgoal**: The current subgoal to be achieved.
- **Current Plan**: The current plan that you need to check and replan if necessary.
- **Reason from Supervisor**: The reason from the Supervisor that indicates whether the plan should continue or be revised.
- **Available Actions**: The available actions to use.
- **History**: The history of actions and observations.

If the reason from Supervisor suggests the plan needs fundamental changes (e.g., wrong approach, invalid parameters, missing prerequisites), set RePlan to true and provide a new plan.

Task: {task_description}
Current Subgoal: {subgoal}
Current Plan: {current_plan}
Reason from Supervisor: {reason}
Available actions:
{admissible_commands}

History: {history}

If replanning, your new plan should use the same format as the current plan:
## Step N
Reasoning: <Your reasoning explaining the step>
Step: <What needs to be accomplished, with specific details>

Your response MUST follow this JSON format:
{{
  "RePlan": <true or false>,
  "Thought": <1-2 sentences explaining your decision, referencing the reason above, or null if RePlan is false>,
  "NewPlan": <new plan using the same format as the current plan, or null if RePlan is false>
}}

Your response:
\end{promptbox}

\begin{promptbox}{DAG Updating Prompt for Supervisor}
You are the High-level Supervisor for a task agent. You are maintaining a DAG indicating the process of the task. A Planner and an Executor are executing the nodes with subgoals in the DAG.
After some nodes have been executed, now the task is still incomplete, you need to check the information you have collected in the history, reason about the task progress, identify what has been accomplished and what still needs to be done, and decide whether and how to update the DAG.

The DAG update process includes three operations:
1. Update node descriptions: Replace placeholders in pending/in_progress nodes with actual values and clearer, more concrete wording.
2. Extend DAG: Add new nodes if needed.
3. Remove nodes: Remove unnecessary or impossible nodes.

You will be provided with:
- **Task Description**: The overall task.
- **Current Step**: The current execution step number.
- **History**: The history of actions and observations.
- **DAG State**: The current state of all nodes in the DAG.
- **Available Actions**: The available actions to use.

Task: {task_description}
Step: {current_step}
History: {history}
DAG State: {dag_state}

Available actions:
{admissible_commands}

Your task:
1. Analyze the current DAG structure: Review ALL nodes, their dependencies, and their current states (pending/in_progress/completed/failed).
2. Check collected information in History: Examine all actions and observations in the history to identify what information has been discovered, what objects have been examined, and what actions have been executed.
3. **Check task completion status:**
   - **Count executable nodes**: Identify how many nodes are currently executable (status is "pending" and all dependencies are completed).
   - **If executable nodes = 0, you MUST add new nodes to continue the task.**
4. **Actively reason about the collected information and task completion:**
   - **Identify what has been done successfully**: Identify which key activities for completing the task have been successfully executed.
   - **Evaluate task completeness**: Assess whether the existing information and completed actions are sufficient to complete the task. Consider:
     * What essential steps from the task description are still missing?
     * Can the execution of pending nodes complete the remaining task steps?
       a) If pending nodes can complete the task, you may not need to update them.
       b) If pending nodes cannot complete the task OR if there are no pending nodes, you MUST add new nodes to ensure the task can be completed.
5. Based on your reasoning from step 3 and 4, identify which nodes need updates and which nodes might be unnecessary or need to be added.
6. Provide your reasoning process, updating decision, and updating details in your output.

Rules for updating:
- Only update descriptions of pending/in_progress nodes.
- If a node is failed, consider adding a new node to deal with the failure.
- DO NOT add nodes with similar descriptions or semantic meaning to any existing nodes with status completed/in_progress/pending.
- DO NOT add nodes with overlapping task scopes. Avoid creating nodes where the same action or manipulation is covered by multiple nodes.
- Make sure new descriptions (in description_updates or new_nodes) are executable with available actions and use specific names from observations when available.
- Make sure new descriptions can be clearly understood by the Planner Module and the Executor Module.
- You should keep that the task could be completed in fewer steps as possible, make sure the DAG structure is efficient and not redundant.
- Your DAG MUST always contain at least one sink node that represents the final action to complete the task.

Your response MUST be valid JSON:
{{
  "thought": "<Your reasoning process here, 2-3 sentences>",
  "need_update": <true or false>,
  "description_updates": [
    {{
      "node_id": "<node_id>",
      "new_description": "<updated description with actual values replacing placeholders, ensuring alignment with the task>"
    }}
  ],
  "new_nodes": [
    {{
      "id": "<optional unique id>",
      "description": "<new subgoal that is executable using available actions and necessary for completing the task>",
      "dependencies": <list of node ids that this node depends on>,
      "dependents": <list of node ids that depend on this node>
    }}
  ],
  "remove_nodes": <list of node ids to remove>
}}

Formatting rules:
- Always provide your reasoning in the "thought" field.
- Only include nodes in "description_updates" if description actually changes - do NOT include unchanged nodes. Keep it empty if no description is updated.

Your response:
\end{promptbox}
\twocolumn

\end{document}